\documentclass[letterpaper, 10 pt, conference]{ieeeconf}  

\IEEEoverridecommandlockouts                              

\usepackage{booktabs}
\usepackage{hyperref}
\usepackage[normalem]{ulem}
\usepackage{ulem}
\usepackage{amsmath, amsfonts, amssymb}
\usepackage{multirow}
\usepackage{graphicx}
\usepackage{subcaption}

\title{\LARGE \bf
Cable Routing and Assembly using Tactile-driven Motion Primitives
}

\author{Achu Wilson$^{*1}$ Helen Jiang$^{*1}$ Wenzhao Lian$^2$ Wenzhen Yuan$^1$ 
\thanks{$^{*}$Authors with equal contribution.}
\thanks{$^{1}$Carnegie Mellon University, PA, USA. \texttt{\{achuw, helenjia, wenzheny\}@andrew.cmu.edu}}
\thanks{$^{2}$Intrinsic Innovation LLC, CA, USA. \texttt{\{wenzhaol\}@google.com}}}

\begin{document}

\maketitle
\thispagestyle{empty}
\pagestyle{empty}

\begin{abstract}

Manipulating cables is challenging for robots because of the infinite degrees of freedom of the cables and frequent occlusion by the gripper and the environment. These challenges are further complicated by the dexterous nature of the operations required for cable routing and assembly, such as weaving and inserting, hampering common solutions with vision-only sensing.
In this paper, we propose to integrate tactile-guided low-level motion control with high-level vision-based task parsing for a challenging task: cable routing and assembly on a reconfigurable task board. Specifically, we build a library of tactile-guided motion primitives using a fingertip GelSight sensor, where each primitive reliably accomplishes an operation such as cable following and weaving. The overall task is inferred via visual perception given a goal configuration image, and then used to generate the primitive sequence. Experiments demonstrate the effectiveness of individual tactile-guided primitives and the integrated end-to-end solution, significantly outperforming the method without tactile sensing. Our reconfigurable task setup and proposed baselines provide a benchmark for future research in cable manipulation. More details and video are presented in \url{https://helennn.github.io/cable-manip/}






\end{abstract}

\section{Introduction}



Robotic manipulation and planning has seen substantial progress in recent years. However, manipulating flexible objects such as cables and wires remains an open problem \cite{grannen2021untangling, pirozzi2018tactile, jin2022robotic, switchgear}. In particular, tasks such as cable assembly require robot systems to be able to effectively track narrow cables and route them in between obstacles. Because the object of interest is thin, malleable, and subject to severe occlusions, this often requires advanced sensing, planning, and control.

While previous efforts in cord manipulation have considered knot tying \cite{grannen2021untangling}, wire insertion \cite{pirozzi2018tactile}, and (most relevant to us) cable routing \cite{jin2022robotic}, they are limited in task complexity and generalizability. Specifically, simplistic operations such as picking, moving, and placing prevent the robot from handling more complex tasks such as weaving parts through slots~\cite{nistcomp}. In addition, the above mentioned methods often cannot recover from failure due to the open-loop nature of the execution~\cite{jin2022robotic}. The main difficulty lies in how to continuously estimate the cable state, particularly when vision alone suffers from severe gripper/object occlusions, and how to generate motion commands accordingly.

\begin{figure}[t]
    \centering
    \includegraphics[width=\linewidth]{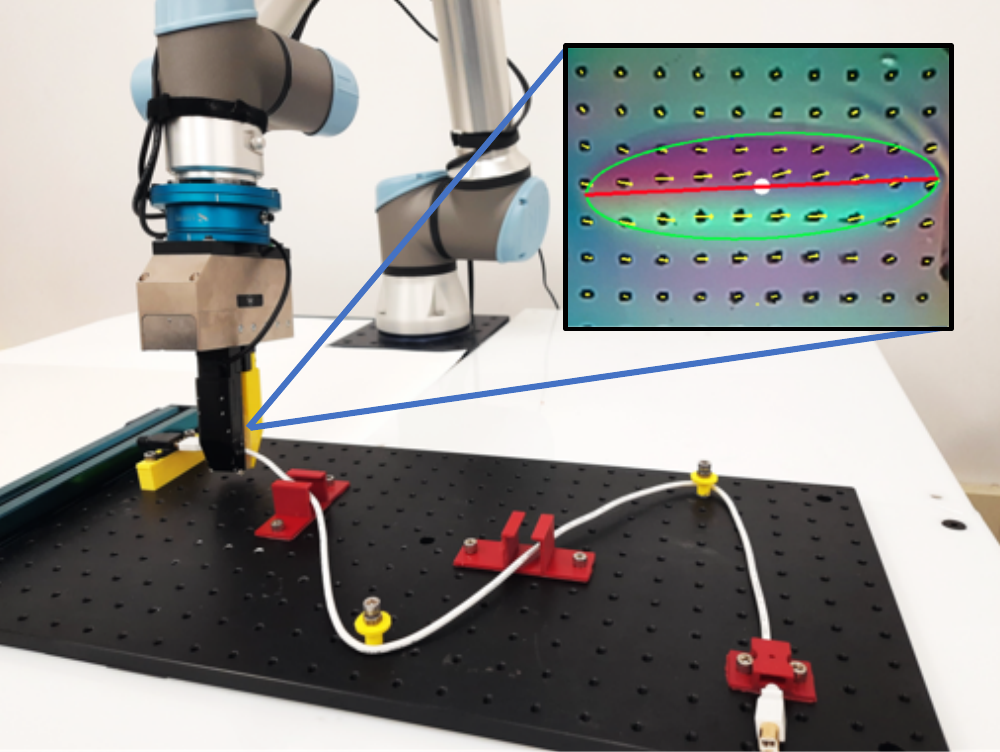}
    \caption{\textbf{Goal-conditioned cable manipulation using visual-tactile perception.} Given an RGBD image of a goal configuration, our system parses the task, generates reference trajectories, and applies a sequence of tactile-guided motion primitives to accomplish a cable routing and assembly task.}
    \label{fig:teaser}
\end{figure}

Tactile-guided manipulation is a promising approach, but has only been demonstrated in a simple cable following task \cite{she2021cable}. In this paper, we consider a full task of cable routing and assembly inspired by the NIST Assembly Task Board 3~\cite{nistcomp}. We design a reconfigurable setup with fixtures that require 4 operations covering common tethered object manipulation, as illustrated in Fig.~\ref{fig:teaser}. These tasks have longer horizons, and require reliable generalization across configurations.

To solve this problem, we propose to use visual perception for task understanding, i.e., a task description file is automatically extracted from a goal configuration image. The parsed task file is then mapped to a sequence of tactile-guided motion primitives. Notably, we leverage tactile sensing to design a library of primitives to continuously estimate the cable state and manipulate it in a closed-loop manner. Concretely, four primitives (cable following, pivoting, weaving, and insertion) are defined, each as a state machine, where the state transitions are governed by tactile perception. State-dependent controllers are then activated sequentially to realize the desired primitive behavior. Because of this modular design where a task is parsed via visual perception into primitives and each tactile-guided primitive is task context independent, our approach is able to generalize across different task board configurations with zero adaptation effort. Experiment results demonstrate the effectiveness and generalizability of the tactile-guided motion primitives, as well as our fully integrated pipeline.

Our contributions are summarized as follows. First, we propose a novel integrated solution for cable routing and assembly, where visual perception enables automatic high-level task parsing. Second, we design a library of tactile-guided motion primitives for low-level motion control to accomplish complex cable operations. Third, we provide baselines with and without tactile sensing on a reconfigurable cable routing and assembly task for future research.

\section{Related Work}


\begin{figure}[t]
    \centering
    \includegraphics[width=\linewidth]{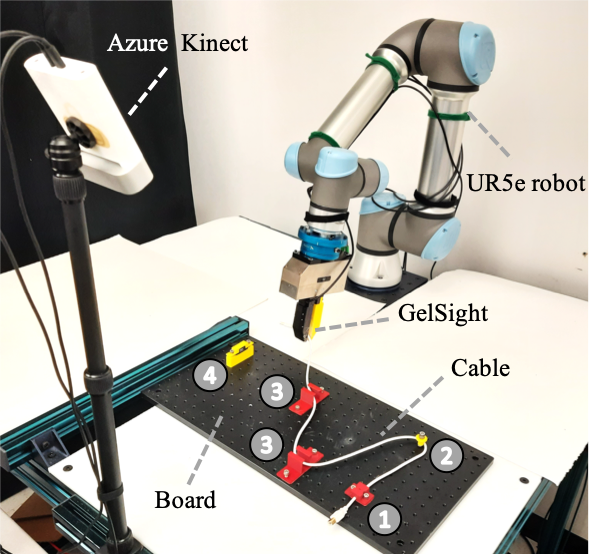}
    \caption{\textbf{System setup and task board configuration.} The task board contains 4 types of fixtures, labeled 1-4 in the figure (1 = start, 2 = pivot, 3 = slot, 4 = USB connector). The cable begins at the start and ends being inserted into the USB connector. In between, the cable may be wrapped around the yellow pivots or woven through red slots. A UR5e robot with a GelSight sensor attached to the gripper is used to manipulate the cable. We also have a Azure Kinect camera overlooking the board to capture the board configuration.}
    \label{fig:setup}
\end{figure}
\begin{figure}[t]
    \centering
    
    \includegraphics[width=\linewidth]{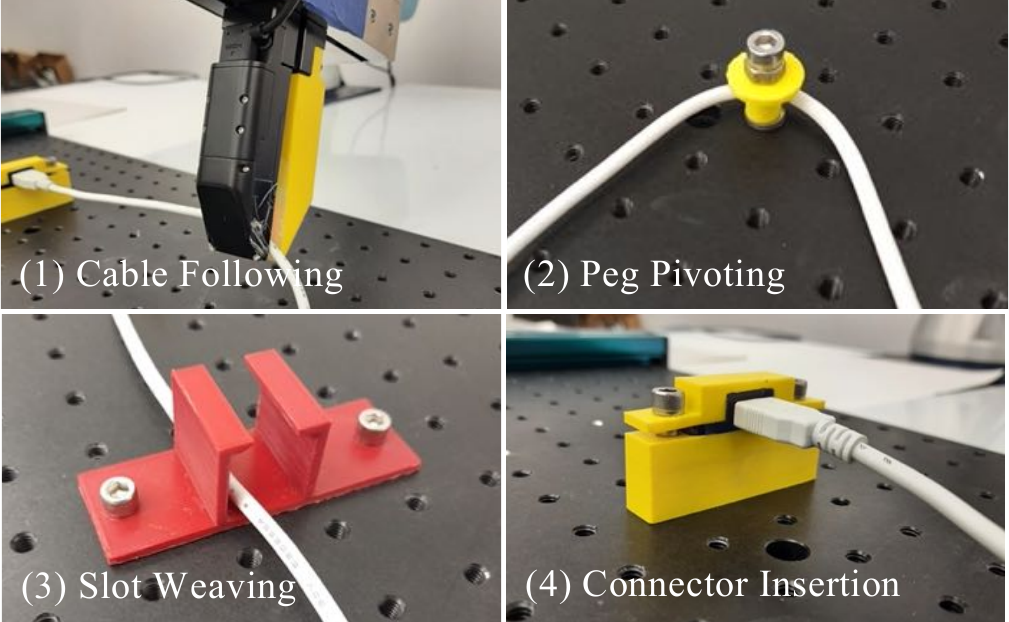}
    \caption{\textbf{Task board operations.} Our task is split into 4 operations: 1) cable following (gripping and following along the cable), 2) peg pivoting (wrapping the cable around the vertical peg), 3) slot weaving (threading the cable through the horizontal slot), and 4) connector insertion (inserting the USB head at the end of the cable into the connector). }
    \label{fig:ref_prim}
\end{figure}

\subsection{Tactile sensing for robotics}
The sense of touch, one of the main reasons that imparts dexterous and fine manipulation skills to human hands, has been inspiring robotics researchers since early days. A multitude of tactile sensing technologies \cite{s18040948} have been developed to aid robot manipulation. This includes measurement of features such as shape, texture, and forces. A recent and increasingly common mode of tactile sensing is optical tactile sensors such as GelSight \cite{yuan2017gelsight} \cite{9811832}. 

GelSight has been used for estimating contact shape and pose of objects~\cite{8794298} and manipulating them~\cite{6943123}. The shear and slip forces on a grasped object can also be estimated \cite{8202149}, which can be used to adjust the grasp \cite{8593528}. \cite{9196976} developed manipulation primitives that exploited tactile sensing for reactive manipulation of tabletop objects. \cite{9341006} used tactile exploration to learn objects' physical properties for dynamic manipulation. \cite{kim2022active} uses tactile sensing to estimate contact locations between a grasped object and the external environment for peg-in-hole insertion. End-to-end learning methods using tactile inputs have been developed for manipulation of small objects \cite{tian2019manipulation}. However, all of the above works focus on rigid object state estimation, while we investigate tasks that require continuous state estimation of deformable objects.

Another line of relevant work is tactile servoing. A neural network based edge detector taking tactile images as input was learned in \cite{8641397}, and was used for contour following. Reinforcement learning approaches for tactile contour following and closing a ziploc bag was developed in \cite{8039205}.
In the spectrum of classical control methods, a tactile servoing controller was designed to track an object edge as well as a cable lying flat on a surface~\cite{li2013control}. In \cite{9353219}, a tactile-based closed-loop velocity controller was developed to regulate the sliding behavior of a grasped object. 
One of the closest work to ours is \cite{she2021cable}, where a GelSight gripper is utilized to enable a closed-loop controller for cable following in free space. In contrast, we tackle a more realistic scenario where the cables rest naturally on a task board, and a more complex task that involves routing and weaving in a constrained workspace.



\begin{figure*}[t]
    \centering
    \includegraphics[width=\linewidth]{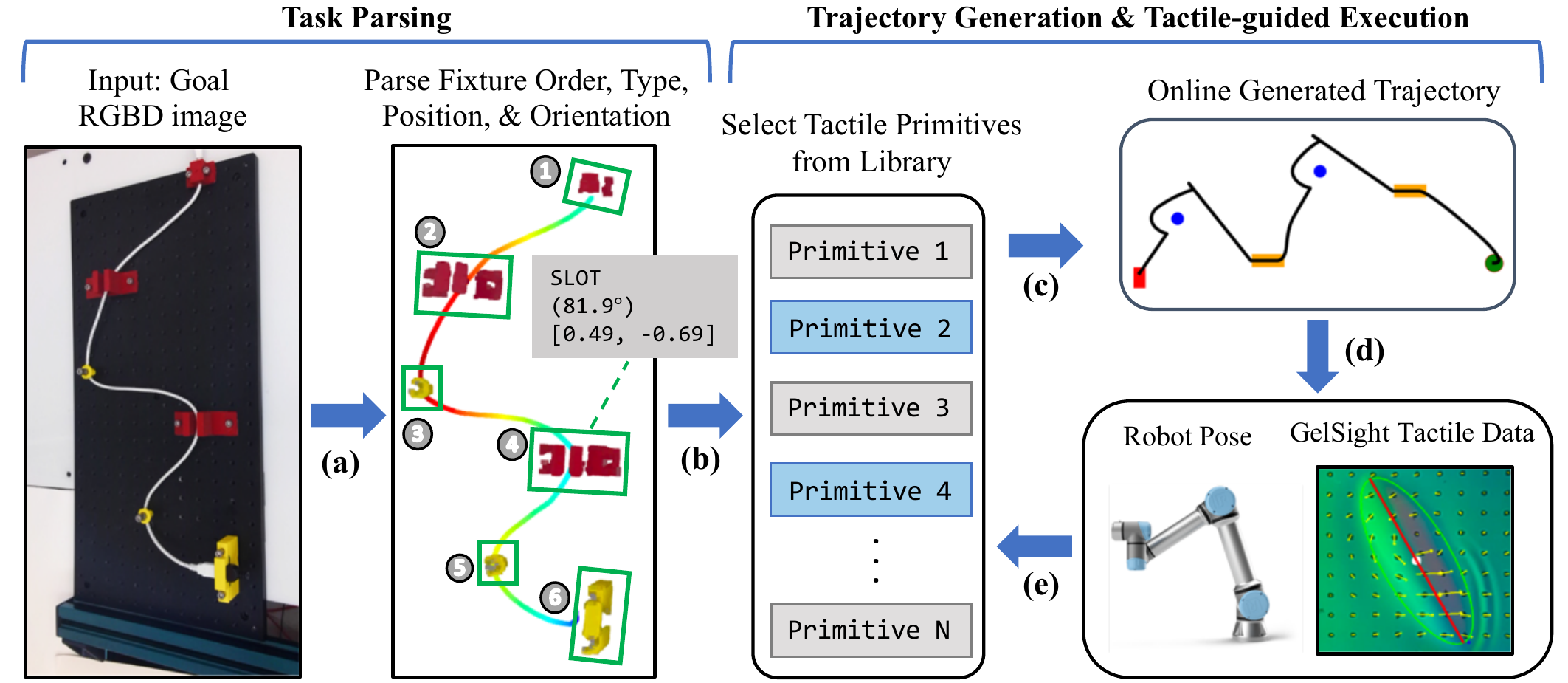}
    \caption{\textbf{Pipeline overview.} We propose an end-to-end framework for cable routing and assembly. 
    \textbf{(a)} Our vision module takes in RGBD images of the goal taskboard with and without the cable to output the task specification. We use color filtering to determine the fixtures' positions and types, Principal Component Analysis (PCA) to determine the orientations and shapes, and Coherent Point Drift (CPD) to detect the cable thus determining the fixture order. \textbf{(b)} The parsed task description (fixture order, type, position and orientation) is mapped to a sequence of parameterized tactile primitives, meanwhile generating an initial reference robot trajectory. \textbf{(c, d, e)} Sequentially, the robot executes each primitive as an individual state machine, where the state transitions are governed by the sensed tactile data. In each state, a parametric trajectory generator is activated to generate the trajectory online, for example, with lines and splines.}

    \label{fig:pipeline}
\end{figure*}


\subsection{Deformable linear object manipulation and assembly}

Manipulating deformable linear objects such as cables and ropes with robots is generally challenging due to the objects' infinite degrees of freedom. Previous works have attempted to design methods ranging from state estimation, representation learning, motion planning, to end-to-end learning.

In an early work~\cite{619320}, the cable deformation due to external forces is estimated using stereo vision, and manipulation techniques are developed to straighten the cable for through-hole insertion. \cite{jin2022robotic} proposes a novel spatial representation between the cable and environment objects for motion planning. In recent years, end-to-end approaches were proposed to learn the deformation model from simulation and manipulate cables to a target shape\cite{Wang_2022}. These works commonly assume the system to be quasistatic and achieves manipulation using repetitive pick and place actions \cite{yan2020self} \cite{nair2017combining}. Few approaches leverage environment contacts when manipulating cables, e.g., leveraging force-torque sensing \cite{feeltension} or visual perception \cite{8851170}. \cite{cableplanning} proposes a task-space planner, which builds a roadmap from predefined tasks and employs a replanning strategy based on a genetic algorithm which executes cable routing using a dual-arm robot. \cite{wireterminalinsertion}  developed a system for insertion of wire-terminal insertion via visuo-tactile methods. \cite{symbolic_state} developed Bayesian state estimation methods to predict symbolic states with predicate classifiers for connector insertion. In this work, we aim to build an entire cable routing and connector insertion system using visual sensing for initial plan generation and then tactile perception to monitor the cable-environment contact state, and adjust the control policy accordingly.








\section{Problem Statement}



\subsection{Task}



We aim to solve the cable manipulation problem inspired by the NIST Assembly Task Board 3\cite{nistcomp}. We consider a taskboard that consists of fixtures such as pegs and channels to route and manipulate the cable to a pre-specified configuration. 
The taskboard consists of 4 different types of fixtures in various configurations: 1) Start fixture, 2) Vertical pegs / pivots, 3) Horizontal slots, and 4) USB connector.
An example of the taskboard with fixtures with its goal configuration is shown in Fig.~\ref{fig:setup}.


\subsection{Operations} 

To move the cable around the fixtures on the board, we divide the entire task into the following subtasks, as illustrated in Fig.~\ref{fig:ref_prim}.

\subsubsection{Cable following} The robot holds the cable and follows along without dropping it.

\subsubsection{Pivoting around vertical pegs} The robot pivots the cable's heading direction by rotating it around the pegs.

\subsubsection{Weaving through horizontal slots} The robot weaves or threads the cable through the horizontal channel slots that ``lock'' the cable inside. 

\subsubsection{Connector insertion} The robot inserts the USB head at the cable's end into the connector fixture.

\section{Method}



The proposed method starts with estimating the configuration of the taskboard and the goal cable configuration from an assembled taskboard. Once the taskboard configuration is estimated, a task description is generated and mapped to a sequence of parameterized tactile primitives. Each primitive is designed as an individual state machine, where the state transitions are governed by the sensed tactile data. The robot executes primitives sequentially until task completion.

\subsection{Perception systems \& task parsing}
The perception system uses RGBD data from a Microsoft Azure Kinect camera overlooking the task board and tactile data generated by a GelSight R1.5 optical tactile sensor.

\subsubsection{Visual perception}
\label{sec:visual_perception}
The visual perception module requires an assembled task board from human demonstration, and infers the goal configuration of the cable.
Given a point cloud from the Kinect camera, we aim to recover the type, position, orientation and ordering of each fixture on the task board. We also track the demonstration cable in the image to parse the order of the fixtures along it. The task parsing section in Fig.~\ref{fig:pipeline} shows an example input-output pair of the visual perception module.

\textbf{Fixture pose estimation:} We first use a point cloud of the peg board with the fixtures but without the cable. We locate all fixtures via simple color filtering and clustering. We apply Principal Component Analysis (PCA) to estimate the type (using the shape, or oblongness) and orientation of each detected fixture. The ``start" and ``connector" fixtures needs their orientation ambiguity to be resolved, so we use the cable heading direction to infer their orientations. Specifically, we take the point from the cable leading up to the given fixture and the current point of the cable the fixture is on to calculate the directional vector.


\textbf{Cable state estimation:} We track the cable in the demonstration board to get the correct order of the routing task. To segment the points corresponding to the white cable, we take all of the white-colored points from the point cloud from the second RGBD image defining the goal cable configuration. The resulting cable point cloud has occlusions created by the fixtures, so we use Reeb Graph \cite{schulman2013tracking} to construct a set of cable nodes. This gives an initialization for Coherent Point Drift (CPD) \cite{myronenko2010point} to complete the cable. With the completed sequence of cable nodes, the order of the fixtures are readily determined by tracing the cable. This cable node sequence also disambiguates the orientation of the ``start" and ``connector" fixtures, as described in the paragraph above. 

Using this visual perception pipeline, the type, location, orientation and order of the fixtures are estimated, which defines the routing and assembly task. This task description then serves as the input for the motion primitives introduced in Section~\ref{sec:primtives}.




\subsubsection{Tactile system}
\label{sec:tactile_system}
We use the tactile reading to estimate the cable state in different stages and adjust the manipulation strategies accordingly. The fingertip GelSight sensor provides tactile images of the contact area and marker displacement information corresponding to the 3-axis forces and in-plane torque on the contact surface.
As introduced in~\cite{yuan2017gelsight}, the change of the color in the GelSight images corresponds to the contact geometry and the motion of the markers in the images indicates the contact force and torque: a ``spreading out'' pattern of the markers' motion indicates normal force, a uniform motion pattern of the markers indicates shear force towards the motion direction, and a spiral pattern indicates an in-plane torque. The magnitude of the marker's motion is approximately linear to the magnitude of the force. 

\begin{figure}[t]
\centering

\smallskip

\includegraphics[width=0.875\linewidth]{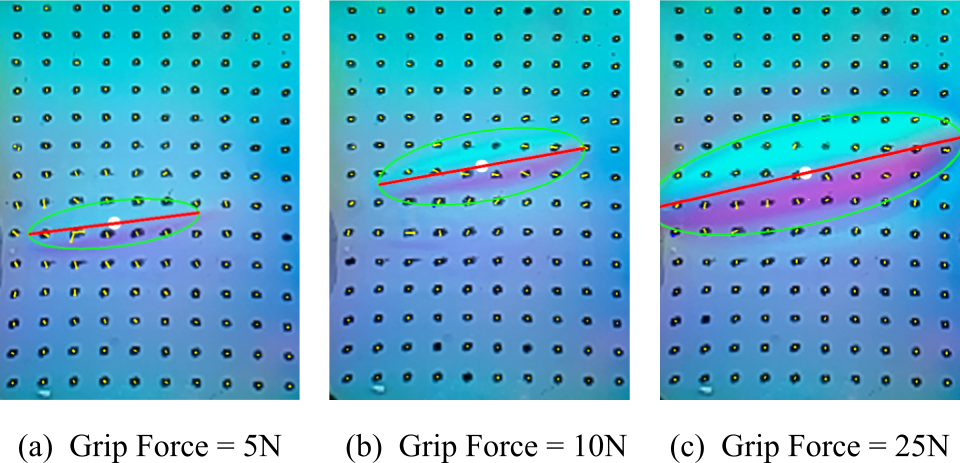} 
 \caption{When holding the cable with different forces, the contact area measured from GelSight is an elliptical shape. The area of the ellipse is roughly linear to the gripping force, and its center and orientation help estimate the cable pose.}
\label{fig:fig_area}
\end{figure}

In this work, we estimate the contact area of the cable based on color and background image subtraction. The contact area of the cable is elliptical in shape, as shown in Fig.~\ref{fig:fig_area}. The area also corresponds to the normal force, which is used to detect the firmness of the cable gripping. We fit an ellipse to the contact area and use its center and major axis to estimate the pose of the cable in hand. This helps the robot to re-center and re-orient the cable in the gripper.

\begin{figure}[t]
\centering

\includegraphics[width=0.875\linewidth]{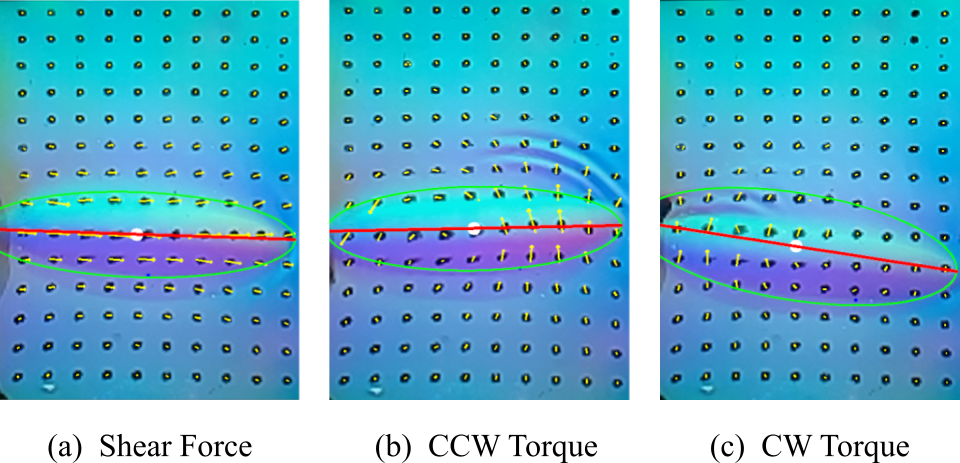} 
 \caption{\textbf{Motion patterns of the GelSight markers.} We use the motion patterns of the GelSight markers to determine contact forces and torques which are used to estimate the cable states in different stages. (a) Shear forces along the cable indicate tension when dragging the cable. (b) A counterclockwise (CCW) torque is generated when the USB cable head hits the connector fixture from the top. (c) A clockwise (CW) torque is produced when the cable rear makes contact with a slot fixture from the top.
  }
\label{fig:fig_contact}
\end{figure}

Force and torque changes from the markers are tracked to identify cable states, such as a slowly increasing shear force along the cable indicating tight cable hold, and fast changing contact force/torque indicating a collision between the cable and the fixture. Fig.~\ref{fig:fig_contact} shows some example tactile images used for state estimation at different stages. 


The GelSight marker magnitudes also serve as input for the hybrid force-position controller used in connector insertion. GelSight data is sampled at 60 Hz with a latency of around 75 ms, while the force-position controller is a cascaded PID controller with an update frequency of 250 Hz. To match the controller frequency, the low frequency GelSight data is linearly interpolated in time. 
To mitigate the destabilizing effects of latency, high-level commands are issued to the controller at a considerably lower frequency than the control loop's operating frequency.

\subsection{Motion primitives and trajectory generation}
\label{sec:primtives}

The vision system infers an ordered list of fixture type, position, and orientation for task specification. This information is utilized to generate a reference robot path, as illustrated in the Online Generated Trajectory subplot in Fig.~\ref{fig:pipeline}, which is divided into sections for each fixture. Motion primitives designed for each task are executed for the corresponding path sections. The robot begins sequentially sets the end position of the previous primitive as the starting position for the next until the task is completed.

The primitives are parameterized and modeled from observing a human performing the task and deriving heuristics.  The primitives are designed as state machines, whose state transitions are triggered by the tactile signals generated by the external forces on the cable, as shown in Fig.~\ref{fig:primitives}. Following are the four primitives:



 \begin{figure}[t]
    \centering
    \includegraphics[width=\linewidth]{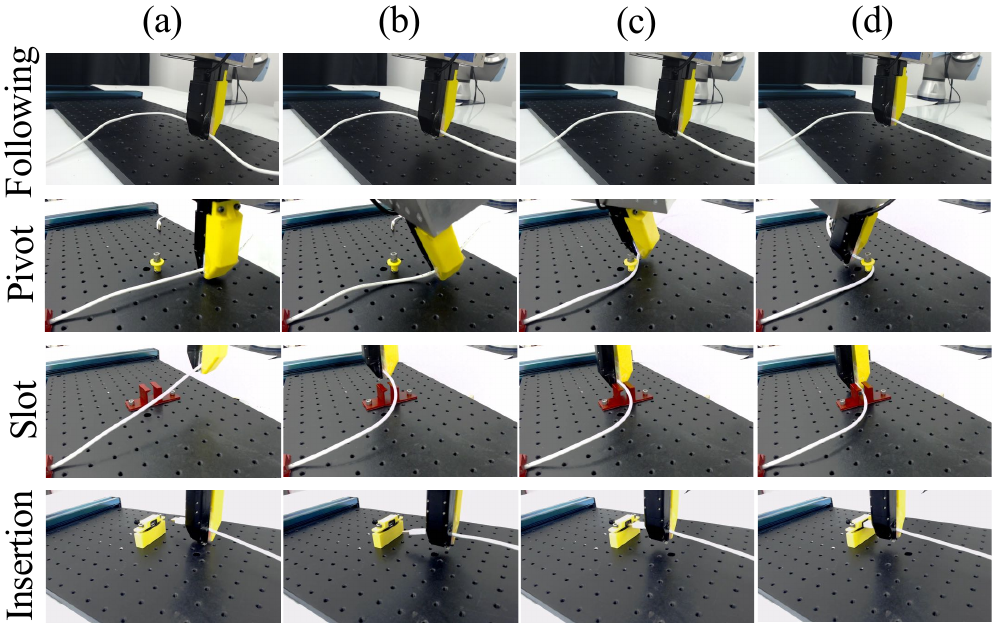}
    \caption{{\bf Motion primitives.} To complete our cable routing and assembly task, we construct a library of motion primitives. The cable following primitive moves the robot along the cable, while pulling it in a desired direction. The pivot primitive wraps the cable around a peg while maintaining cable tension. The slot primitive inserts the cable into the slot and ``locks it" in place. The insertion primitive aligns and inserts the USB connector into the socket.}
    \label{fig:primitives}
\end{figure}


\subsubsection{Cable following}
\label{sec: cable_following}
The robot performs this primitive to guide the cable from one fixture to another.
As illustrated in the first row of Fig.~\ref{fig:primitives}, the primitive executes the following state sequence in a loop: i) the gripper slowly closes until the gripping force detected by GelSight exceeds a threshold; ii) the robot pulls the cable until the sensed shear force exceeds a threshold, indicating the cable tension; iii) the gripper slowly opens until the gripping force falls below a threshold; iv)  based on the detected cable pose in the gripper, the robot slides along the cable while keeping the cable centered.

\subsubsection{Pivoting around pegs}

The pivoting primitive changes the direction of the cable to make it pivot around a vertical peg, as shown in the second row in Fig.~\ref{fig:primitives}. 
This primitive starts with following along the cable until a waypoint, determined by the peg position. The robot tilts backwards and moves down until the tactile signals indicate the cable is in contact with the taskboard, as illustrated in Fig.~\ref{fig:primitives} top row, plot (b) . These tactile signals looks similar to the clockwise torque signals in Fig.~\ref{fig:fig_contact} (c), ensuring that the cable gets tucked under the pivoting peg. The robot then moves in a circular trajectory centered at the initial position, and continues until the cable makes contact with the peg as in plot (c). 
After establishing contact between the cable and peg, a new circular trajectory is generated using the peg's position as the center.  

\subsubsection{Weaving through slots}
The weaving primitive is used to guide the cable through horizontal slots, as shown in the third row of Fig.~\ref{fig:primitives}.
It is parameterized by the slot position and orientation, which are used to generate a reference trajectory to place the cable into the slot from the top. If the cable is not aligned with the slot during the downward motion, as indicated by a sudden increase of the in-plane torque from the GelSight image, the robot executes a horizontal wiggling action while moving down. It is continued until the in-plane torque disappears, signaling that the collision is resolved and the cable is aligned with the slot. 



\subsubsection{Connector insertion}

This primitive conducts tethered USB connector insertion given the socket position and orientation, as shown in the bottom row of Fig.~\ref{fig:primitives}. 
%
The primitive starts from a grasp on the cable near the connector, which enables a certain amount of freedom for the connector to bend when hitting the rigid side of the socket fixture. When the connector contacts the socket fixture, the GelSight sensor detects a normal force along the cable. We then command the robot using a hybrid force-position controller to rotate around the normal axis while keeping the normal force. The motion enables the connector to move for a small distance around the initial contact location. When the connector falls into the socket, there will be a sudden drop in the normal force on the cable and a strong torque that stops the cable from continuing rotation. Both the change of force and torque can be detected by GelSight, and we will make the robot stop this ``exploration'' procedure and push forward to insert the connector into the socket. In some cases, if the initial contact point is too far away from the socket, the rotary exploration motion will not get the connector inside the socket. We will then make the robot retreat and start the exploration from another randomly-sampled location near the socket.

\section{Experiments}

In this section, we evaluate the performance of our cable routing and assembly pipeline. Specifically, we investigate the benefits of tactile sensing in cable manipulation tasks and the reliability of our designed tactile primitives.



\subsection{Robot setup}

Our full setup is shown in Fig.~\ref{fig:setup}. We use a UR5e 6-DoF robot arm  switching between position control and hybrid force-position control modes. 
To grasp the cable, we use a Weiss Robotics WSG-50 gripper, which operates in an indirect force control. Mounted on the gripper is a GelSight R1.5 sensor along with a custom designed and 3D printed finger which has design features to prevent accidental cable drops. These design features include an inner surface with an elastomer pad with adequate friction, a shape which bulges slightly towards the tip, and a beak on one side which helps picking up thin cables from the flat surface and holds it from falling down. We also use a Microsoft Azure Kinect camera to capture the entire taskboard and acquire RGBD data for task parsing. 


\begin{table}[t]
\centering
\resizebox{\linewidth}{!}{ 
\begin{tabular}{lcccc}
\toprule \addlinespace[0.15cm]
Primitive & \begin{tabular}[c]{@{}c@{}}Cable\\Following\end{tabular} & \begin{tabular}[c]{@{}c@{}}Peg\\Pivoting\end{tabular} & \begin{tabular}[c]{@{}c@{}}Slot\\Weaving\end{tabular} & \begin{tabular}[c]{@{}c@{}}Connector\\Insertion\end{tabular} \\
\midrule
w/o Tactile (Baseline) & 43/50 & 28/50 & 32/50 & 0/50 \\
w/ Tactile (Ours) & \textbf{50/50} & \textbf{50/50} & \textbf{50/50} & \textbf{48/50}\\

\bottomrule
\end{tabular}
}

{\small}
\caption{\textbf{Success rate of individual primitives.} For each primitive, we calculate the success rate with and without tactile sensing. 
\label{tab:successrate}} 
\end{table}

\subsection{Motion primitives with and without tactile sensing}

We demonstrate the robustness of the tactile-guided motion primitives by comparing the success rate against a baseline without tactile sensing. The baseline applies the same trajectory which is generated analytically and does not use tactile sensing for adapting it.

We run 50 trials for each primitive (see Fig.~\ref{fig:primitives}) with varying fixture positions, and present the results in Table~\ref{tab:successrate}.A trial is considered success if the robot is able to properly perform the corresponding primitive on that fixture. For all four primitives, our tactile-guided approach significantly outperforms the baseline without tactile sensing. The connector insertion primitive shows this the most clearly: the baseline has a success rate of 0\% because the robot cannot align the orientation of the USB head with the connector, whereas with the tactile sensing, the controller can reorient the connector pose to align with the socket. The other baseline primitives fail due to the inability to estimate contacts and maintain tension in the cable.

\begin{table}[t]
\centering

\begin{tabular}{@{}lcc@{}}
\toprule \addlinespace[0.15cm]
          & Success Rate & Failure Modes     \\ \midrule
GT + Tactile (Ceiling) & 53/60 & A(3), B(2), C(2), D(0) \\
GT + No Tactile (Baseline) & 0/60 & A(0), B(2), C(10), D(48) \\
Vision + Tactile (Ours)    &   \textbf{51/60}   & A(4), B(3), C(2), D(0) \\ 
\bottomrule
\end{tabular}
{\small
\caption{\textbf{Success rate of entire pipeline.} We summarize the success rate of the whole task across 6 different board configurations with 10 trials each. Failure Modes: A (Connector insertion failed), B (Cable got stuck), C (Cable escaped from previous fixture), D (Accumulated error in motion)}\label{tab:full_traj}}

\end{table}
\begin{figure}[t]
    \centering

    \includegraphics[width=\linewidth]{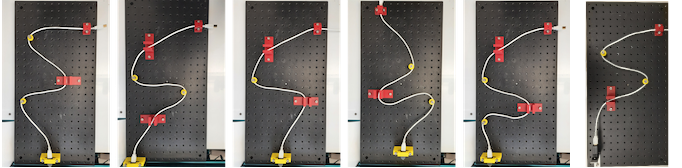}
    \caption{\textbf{Taskboard configurations.} We evaluate cable routing and assembly on each of these 6 boards 10 times.}
    \label{fig:boardconfigs}
\end{figure}

\subsection{End-to-end cable routing and assembly}
We analyze the reliability of our full pipeline over various board configurations. First to evaluate the performance of our vision pipeline described in Section~\ref{sec:visual_perception}, we compare our approach (Vision + Tactile) against an oracle baseline that uses the ground truth (i.e. human-specified) board configuration (GT + Tactile). We consider this to be an upper bound on performance.
To evaluate the effectiveness of using tactile sensing, we experimented with a baseline without tactile sensing but using the ground truth board configuration (GT + No Tactile).

We compare the success rate and failure modes by evaluating each approach across 6 different board configurations (Fig.~\ref{fig:boardconfigs}), with 10 trials each, summarized in Table~\ref{tab:full_traj}. 
We label the trial as a success if the entire task was completed.
The oracle which uses ground truth board configurations (53/60) performs only slightly better than our Vision + Tactile pipeline (51/60), suggesting that our vision pipeline can effectively parse the board task configuration. 
We find that the baseline without tactile sensing fails catastrophically due to the accumulated motion error, typically after the second or third fixture.  While individual primitives could be completed without tactile sensing, although with limited success, as shown in Table~\ref{tab:successrate},  we find that chaining the primitives reliably is difficult. 
This happens due to the lack of tactile feedback which corrects the trajectory while following the parsed path. The robot may also experience failures beyond its direct control, such as cable entanglement in fixtures or cable detachment from previously completed fixtures. 
During complex operations such as pivoting, weaving, and insertion, tactile signals are crucial for inferring the cable-environment interaction state to online adapt the trajectory. As summarized in the failure modes, our method occasionally fails because of the cable getting stuck or slack, and connector insertion remains the most challenging operation. We postulate such ``global'' state changes cannot be sufficiently captured by ``local'' tactile signals. Thus we plan to fuse the global visual sensing with local tactile sensing in the future to further improve the reliability.



\section{Conclusion and Future Work}
This paper studies a long-horizon task that involves cable routing and tethered object assembly. Particularly, we propose a novel integrated pipeline where a task is parsed via visual perception and trajectories are planned and executed in a closed-loop manner with tactile-guided motion primitives. We compare the designed method and other baselines on a reconfigurable task board, which may serve as a benchmark for future research. Experiment results indicate the necessity of tactile sensing in such tasks involving dexterous operations with cables in a realistic environment.

One key limitation of our work is that the primitives are hand-designed rather than learned from data, thus limiting their generalizability. It is interesting to create more diverse and powerful primitives leveraging imitation learning and reinforcement learning. Another limitation is the visual and tactile perception being used at different stages. In future work, we plan to fuse the sensor modalities, joining global and local information, and learning task-relevant state representations.




\addtolength{\textheight}{-4 cm}   







\bibliography{bibliography.bib}
\bibliographystyle{IEEEtran}

\end{document}